\documentclass[sigconf,screen]{acmart}

\usepackage{diagbox}
\usepackage{times}
\usepackage{epsfig}
\usepackage{graphicx}
\usepackage{graphics}
\usepackage{amsmath}
\usepackage{booktabs}
\usepackage{pgfplots}
\usepackage{multicol}
\usepackage{multirow}
\usepackage{subfigure}
\usepackage{tikz}
\usepackage{appendix}
\usepackage{comment}
\usepackage{emptypage}
\usepackage{geometry}
\AtBeginDocument{%
  \providecommand\BibTeX{{%
    \normalfont B\kern-0.5em{\scshape i\kern-0.25em b}\kern-0.8em\TeX}}}

\author{Zihao Wang}
\authornote{Equal Contribution}
\affiliation{%
  \institution{Beijing University of Technology}
  \city{Beijing}
  \country{China}
  }
\email{Rex.Wangzihao@gmail.com}

\author{Yiming Huang}
\authornotemark[1]
\affiliation{%
  \institution{Beijing University of Technology}
  \city{Beijing}
  \country{China}
  }
\email{huangyiming2002@126.com}

\author{Ziyu Zhou}
\authornotemark[1]
\affiliation{%
  \institution{Beijing University of Technology}
  \city{Beijing}
  \country{China}
  }
\email{ziyuzhou30@gmail.com}

\usepackage{fancyhdr}
\renewcommand\footnotetextcopyrightpermission[1]{}

\usepackage[small,compact]{titlesec}

\pgfplotsset{compat=1.18}
\begin{document}

\setlength{\abovedisplayskip}{1pt} 
\setlength{\belowdisplayskip}{1pt}
\setlength{\floatsep}{2pt plus 2pt minus 2pt}
\setlength{\textfloatsep}{2pt plus 2pt minus 2pt}
\setlength{\intextsep}{2pt plus 2pt minus 2pt}

\title{CoC-GAN: Employing Context Cluster for Unveiling a New Pathway in Image Generation}


\begin{abstract}
Image generation tasks are traditionally undertaken using Convolutional Neural Networks (CNN) or Transformer architectures for feature aggregating and dispatching. Despite the frequent application of convolution and attention structures, these structures are not fundamentally required to solve the problem of instability and the lack of interpretability in image generation. In this paper, we propose a unique image generation process premised on the perspective of converting images into a set of point clouds. In other words, we interpret an image as a set of points. As such, our methodology leverages simple clustering methods named Context Clustering (CoC) to generate images from unordered point sets, which defies the convention of using convolution or attention mechanisms. Hence, we exclusively depend on this clustering technique, combined with the multi-layer perceptron (MLP) in a generative model. Furthermore, we implement the integration of a module termed the 'Point Increaser' for the model. This module is just an MLP tasked with generating additional points for clustering, which are subsequently integrated within the paradigm of the Generative Adversarial Network (GAN). We introduce this model with the novel structure as the Context Clustering Generative Adversarial Network (CoC-GAN), which offers a distinctive viewpoint in the domain of feature aggregating and dispatching. Empirical evaluations affirm that our CoC-GAN, devoid of convolution and attention mechanisms, exhibits outstanding performance. Its interpretability, endowed by the CoC module, also allows for visualization in our experiments. The promising results underscore the feasibility of our method and thus warrant future investigations of applying Context Clustering to more novel and interpretable image generation.
\end{abstract}
\ccsdesc[500]{Computing methodologies~Computer vision}

\keywords{Context Cluster, Generative Adversarial Network}
\maketitle
\section{Introduction}
In recent developments, a pioneering approach known as Context Cluster (CoC)~\cite{CoC} has emerged, challenging the prevalent use of convolutional networks and attention mechanisms, which have traditionally dominated the field of Computer Vision (CV). This innovative methodology, introduced by Ma et al., underscores the critical role of image interpretation in the feature extraction process. In their work, they operationalize their philosophy by employing similarity measures to cluster sets of points.

Despite their notable success in various dense-to-sparse prediction tasks, such as image classification, an intriguing question arises: can this intuitive method tackle the inverse problem, specifically in the domain of generative tasks? Current challenges in the realm of Generative Adversarial Networks (GAN)~\cite{GAN}, including training instability and weak interpretability, persist unresolved.
To address these concerns, we introduce a novel GAN architecture, underpinned by CoC, to evaluate its feasibility and potential impact on the field.

In the pursuit of refining the paradigm for implementing GANs, the goal has consistently been the evolution of superior generalization capabilities for image data. Goodfellow et al. introduced the intriguing concept of GANs that intuitively learn the prior for sampling from a statistical perspective. Models such as DCGAN~\cite{dcgan} and TransGAN~\cite{gransgan} were among the pioneers in applying convolution and transformer paradigms within GANs, leveraging the inherent image-friendly nature of convolutional networks and the scale-friendly attributes of attention-based methods, respectively.

\begin{figure}[!htbp]
    \centering
    \includegraphics[width=0.6\linewidth]{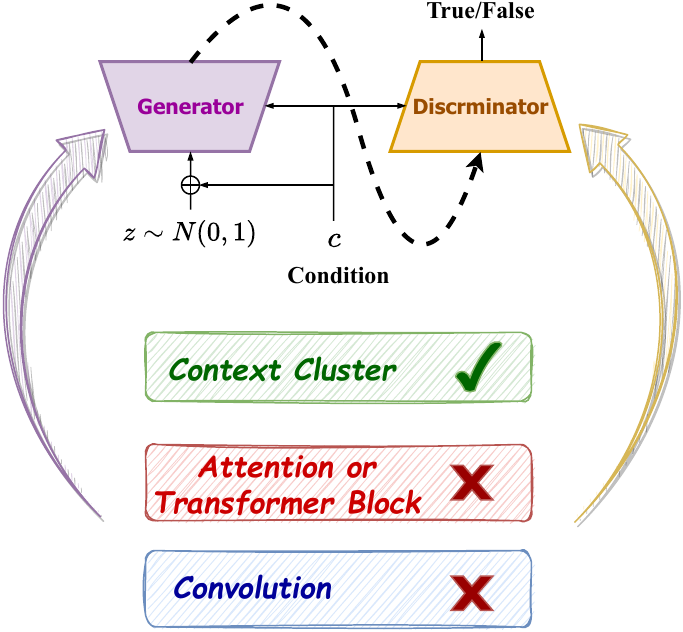}
    \caption{Our motivation: Make an convolution \& attention-free GAN by Context Cluster.}
    \label{fig:mov}
\end{figure}

Despite the advancements, these methods do not fundamentally address the inherent instability of training GANs, and issues surrounding weak interpretability persist, particularly when a high level of reliability is required. This paper introduces a unique GAN architecture, named CoC-GAN, which eschews the conventional use of convolution and attention mechanisms found in earlier GANs. Our motivation is illustrated in Fig.~\ref{fig:mov}. For the discriminator in our proposed GAN, we straightforwardly apply the CoC. The central concept involves viewing the generative process as an augmentative operation on sets of points at different stages, facilitated by the Point Increaser module we have designed. This can be viewed as the functional opposite of the point reduction operation in the original CoC. We undertake empirical experiments to validate our method, encompassing both unconditional and conditional image synthesis.

The contributions of this paper are three-folded:
\begin{itemize}
  \item Our study introduces CoC-GAN, a Context Cluster-based architecture for GAN image synthesis, eliminating the need for convolution and attention mechanisms.
  \item We propose a 'Point Increaser' block that utilizes a simplified clustering method and Multi-Layer Perceptron (MLP) for image generation from unordered point sets.
  \item We validate that the context clustering method significantly enhances the interpretability of image generation process.
\end{itemize}

\section{Related Works}
\subsection{Paradigms in Computer Vision} 
Deep learning has profoundly changed the algorithm and benchmark in the Computer Vision area. To the best of our knowledge, convolution nets (ConvNets) and attention mechanisms make the main contribution to this progress. ConvNets have shown their ability to capture translation invariance of image data from LeNet~\cite{LeNet}. Classic works in the deep learning era like AlexNet~\cite{alexnet}, ResNet~\cite{resnet} etc. shows superiority of fewer parameters by its local parameter sharing property compared to the simple multi-layer perceptron. The drawback of ConvNets is the ability to generalize knowledge in the large-scale dataset. The emergence of ViT~\cite{vit} marked attention-based Transformers gradually solve the problem of inductive bias in ConvNets, due to its bigger capacity and adaptive global interaction. Therefore more methods like SwinTransformer~\cite{swintrans} explore utilizing both advantages of ConvNets and attention. However, researchers are also curious about methods other than these two mainstream ways. Before occurence of CoC, ViG~\cite{vig} and MLP-mixer~\cite{mipmixer} respectively use graph structure and pure multi-layer perceptron in a well-organized structure to realize this goal. And CoC uses clustering ideas to aggregate similar points in the image for the interaction of pixels and super-pixels , which fuse the information for simple downstream tasks. Our work is further exploring its traits in harder generative tasks.

\subsection{Architecture in GAN} 
Adversarial method GAN is the most wide-used method for generative tasks like image synthesis tasks, which simply let adversaration between Generator and Discriminator learn the prior distribution. The first practice by Goodfellow et al. adopts the form of MLP, which shows its ability in small datasets like MNIST~\cite{MNIST}. To Enhance the specific image generation ability, DC-GAN~\cite{dcgan} symbolizes the placement by better ConvNets, the significant effect was shown in different benchmarks. TransGAN~\cite{gransgan} is the first work to expand the strong fitting ability of the attention-based transformer for GANs. Same to the application combining ConvNets and attention in a wide CV area, GigaGAN~\cite{gigagan} also realizes this. In comparison, our novelty is that we are the first to attempt to apply Context clusting (CoC) method in GANs for generative tasks.

\section{Methodology}
\subsection{Overview}
\begin{figure}[!htbp]
    \centering
    \includegraphics[width=0.7\linewidth]{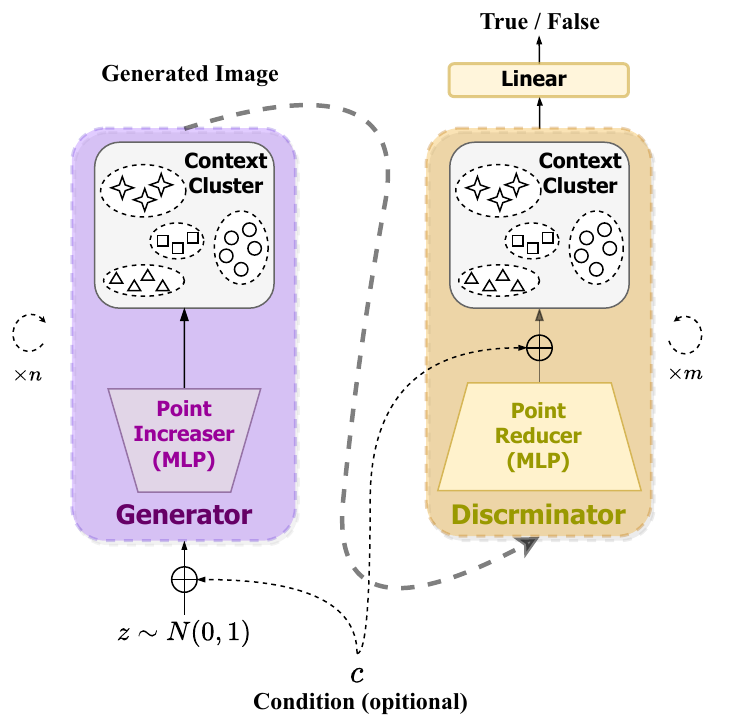}
    \caption{Architecture of our CoC-GAN}
    \label{fig:arch}
\end{figure}

Compared to the original CoC model, the most two prominent characteristics are that we propose the Point Increaser module to generate dense points and we employ the CoC module in both generator and discriminator of GAN. This convolution \& attention-free architecture is demonstrated in Fig.~\ref{fig:arch}.

\subsection{Point Increaser}
In contrast to the downsampling effects produced by the Point Reducer Module in the original CoC model, our task necessitates the creation of a module with an antithetical function for generation tasks. It is evident that we can merely invert the Point Reducer by applying an analogous linear layer for upsampling. For the sake of simplicity and stable implementation, distinct input channels are transformed in a piecemeal manner via a linear layer specifically designed for upsampling.

\subsection{CoC Module}
We utilize original Context Clustering modules within our architecture, where the input is a set of feature points denoted by $\mathbf{P \in R^{n \times d}}$. Initially, we project $\mathbf{P}$ into $\mathbf{P_s}$. Subsequently, we propose $c$ centers in the space by computing the average of the nearest k points. Following this, based on the calculated pairwise cosine similarity matrix $\mathbf{S \in R_{c \times n}}$, we allocate each point to the cluster represented by the centermost similar to it. The aforementioned procedures constitute the clustering stage, setting the groundwork for the subsequent feature dispatching and aggregation stage.

In the feature aggregation stage, all points are adaptively aggregated based on their similarity. Given $\mathbf{m}$ points in a cluster and their similarity to the center of this cluster, we transform them into a value space to obtain $\mathbf{P_v \in R^{m \times d'}}$ where $\mathbf{d'}$ denotes the dimension. The aggregated feature is generated through the following procedure:
\begin{align}
g=\frac{1}{C\left(v_{c}+\sum_{i=1}^{m} \operatorname{sig}\left(\alpha s_{i}+\beta\right) * v_{i}\right)}, \text{ s.t. } C=1+\sum_{i=1}^{m} \operatorname{sig}\left(\alpha s_{i}+\beta\right) . 
\end{align}
$\mathbf{v_c}$ is the center in value space, $\mathbf{\alpha, \beta}$ are the learnable parameters for controling similarity and $\mathbf{\operatorname{sig}(\cdot)}$ is the sigmoid function which scale the similarity in $\mathbf{(0, 1)}$, and $v_i$ refers to i-th point in $\mathbf{P_v}$. 

Regarding the feature dispatching stage, the aggregated feature $\mathbf{g}$ is dispatched to every point in the cluster by the similarity. To be specific, each $\mathbf{p_i}$ is updated by following equation:
\begin{align}
p_{i}^{\prime}=p_{i}+\mathrm{FC}\left(\operatorname{sig}\left(\alpha s_{i}+\beta\right) * g\right)
\end{align}
The fully-connected (FC) layer takes the transform from $\mathbf{d'}$ to $\mathbf{d}$ into account.

The original CoC architecture also incorporates multi-head computation, which is used to transform $\mathbf{P_v}$ into $\mathbf{P_s}$ across $\mathbf{h}$ heads. A Fully-Connected (FC) layer is deployed to fuse the concatenated outcomes derived from each head.

\begin{table*}[!htbp]
\caption{Hyperparameter we select for concrete CoC-GAN in our tasks. The left part of the "Points" column represents the point numbers of the generator and the right part represents the point numbers of the discriminator. The output/input channel(s) of generator/discriminator is/are determined by different datasets.}
\centering
\renewcommand\arraystretch{0.8}
\begin{tabular}{|c|c|c|c|c|c|c|c|}
\hline
\multirow{3}{*}{Stage S1} & Points & Block & \multicolumn{2}{|c|}{CoC-GAN: Generator} & \multicolumn{2}{|c|}{CoC-GAN: Discrminator} \\
\cline{2-7}
& 1 * 1/28 * 28 & Point Increaser/Reducer & $\left\{\begin{array}{c}sample\_r=2 , dim=128\end{array}\right\}$ & $\times 1$ & $\left\{\begin{array}{c} sample\_r=2 , dim=1 (or 3)\end{array}\right\}$ & $\times1$ \\
\cline{2-7}
& 2 * 2/14 * 14 & Context Cluster Blocks & $\left\{\begin{array}{c} heads=4 , head\_dim=16 \\ mlp\_r.=4 , dim=64\end{array}\right\}$ & $\times 2$ & $\left\{\begin{array}{c} heads=4 , head\_dim=16 \\ mlp\_r.=4 , dim=32\end{array}\right\}$ & $\times 2$ \\
\hline
\multirow{2}{*}{Stage S2} 
& 2 * 2/14 * 14 & Point Increaser/Reducer & $\left\{\begin{array}{c}sample\_r=2 , dim=64\end{array}\right\}$ & $\times 1$ & $\left\{\begin{array}{c}sample\_r=2 , dim=32\end{array}\right\}$ & $\times 1$ \\
\cline{2-7}
& 4 * 4/7 * 7 & Context Cluster Blocks & $\left\{\begin{array}{c} heads=4 , head\_dim=16 \\ mlp\_r.=8 , dim=32\end{array}\right\}$ & $\times 2$ & $\left\{\begin{array}{c} heads=4 , head\_dim=16 \\ mlp\_r.=8 , dim=64\end{array}\right\}$ & $\times 2$ \\
\hline
\multirow{2}{*}{Stage S3} 
& 4 * 4/7 * 7 & Point Increaser/Reducer & $\left\{\begin{array}{c} sample\_r=7 , dim=32\end{array}\right\}$ & $\times 1$ & $\left\{\begin{array}{c} sample\_r=7 , dim=64\end{array}\right\}$ & $\times1$ \\
\cline{2-7}
& 28 * 28/ 1 * 1& Context Cluster Blocks & $\left\{\begin{array}{c} heads=4 , head\_dim=16 \\ mlp\_r.=4 , dim=1 (or 3)\end{array}\right\}$ & $\times 1$ & $\left\{\begin{array}{c} heads=4 , head\_dim=16 \\ mlp\_r.=4 , dim=128\end{array}\right\}$ & $\times 1$ \\
\hline
\end{tabular}
\label{Tab:hyper}
\end{table*}

Drawing from the work of Xu et al., as well as our own implementation, the CoC module enhances the global feature through clustering. In the case of downsampling, learnable parameters are trained to proficiently capture useful global features during aggregation and dimension-increasing transformations. However, our generator is designed to validate this concept, or its variation, in an upsampling scenario.

\subsection{Whole Model in Unconditional/Conditional Generation Task}
As illustrated in Fig.~\ref{fig:arch}, our novel approach employs CoC to create a convolution and attention-free architecture for GANs. Hence, we mimic the first GAN~\cite{GAN} for unconditional generation tasks, which involves the use of a Point Increaser in the generator to incrementally increase the number of points from a random seed. The discriminator, in this setup, is a straightforward implementation of the CoC model.
The conditional case presents a more intricate scenario. We adhere to the design of GAN-INT-CLS~\cite{GANINTCLS} to ensure simplicity and conciseness in our approach. In this setup, half of the random seeds in the unconditional variant are replaced with features extracted from the condition. These extracted features also serve as the conditional input for the discriminator. Fully-Connected (FC) layers are employed to adjust the dimensionality of the condition, ensuring that it matches the input.

\begin{table*}[!htbp]

\caption{The comparison between our CoC-GAN and basic GAN model.}
\centering
\tabcolsep=0.1cm
\renewcommand\arraystretch{0.85}
\begin{tabular}{c ccc ccc ccc} 
\toprule
\large Dataset & \multicolumn{3}{c}{MNIST} & \multicolumn{3}{c}{FashionMNIST} & \multicolumn{3}{c}{CryptoPunks} \\ \cmidrule(lr){2-4} \cmidrule(lr){5-7} \cmidrule(lr){8-10}
\large Metric & IS$\uparrow$ & FID$\downarrow$  & KID$\downarrow$ & IS$\uparrow$ & FID$\downarrow$ & KID$\downarrow$ & IS$\uparrow$ & FID$\downarrow$ & KID$\downarrow$  \\
\hline
DCGAN & 2.01 $\pm$ 0.076 & 46.36 & 0.065 $\pm$ 0.001 & 3.26 $\pm$ 0.34 &51.37  & 0.355 $\pm$ 0.006 & 2.58 $\pm$ 0.085 & 41.55 &  0.318 $\pm$ 0.002 \\
WGAN & 2.31 $\pm$ 0.081& 26.53 & 0.020 $\pm$ 0.001& 3.77 $\pm$ 0.18& 39.66& 0.273 $\pm$ 0.003& 2.49 $\pm$ 0.091&  38.52& 0.032 $\pm$ 0.002 \\
WGAN-GP & 2.39 $\pm$ 0.024&  \textbf{25.66}& 0.024 $\pm$ 0.001& 3.74 $\pm$ 0.04& 33.95& 0.241 $\pm$ 0.002& 2.43 $\pm$ 0.085& 36.12& 0.291 $\pm$ 0.002 \\
\textbf{CoC-GAN (unconditional)} & 2.89 $\pm$ 0.033& 28.31 & 0.018 $\pm$ 0.002& 3.88 $\pm$ 0.19& 31.53& 0.226 $\pm$ 0.003& 2.81 $\pm$ 0.019& 30.99 & 0.281 $\pm$ 0.029 \\ 
\textbf{CoC-GAN (conditional)} & \textbf{2.92 $\pm$ 0.021}& 26.47 & \textbf{0.017 $\pm$ 0.001}& \textbf{3.90 $\pm$ 0.20}& \textbf{30.95}& \textbf{0.215 $\pm$ 0.003}& \textbf{2.89 $\pm$ 0.022}& \textbf{29.81} & \textbf{0.263 $\pm$ 0.027} \\ \bottomrule
\end{tabular}
\label{tab:result}
\end{table*}

\subsection{Training objective}
Loss is also a significant factor in validating. We regard two general kinds of loss for learning: hard ones and easy ones. To be representative, we select the most simple loss in ~\cite{GAN} called min-max game as the former:
\label{eq:loss1}
\begin{align}
\min _G \max _D V(D, G) = & \mathbb{E}_{\boldsymbol{x} \sim p_{\text{data }}(\boldsymbol{x})}[\log D(\boldsymbol{x})] +  \notag \\ & \mathbb{E}_{\boldsymbol{z} \sim p_z(\boldsymbol{z})}[\log (1-D(G(\boldsymbol{z})))]
\end{align}
and its conditional version is:
\begin{align}
\label{eq:loss2}
\min _G \max _D V(D, G)=  & \mathbb{E}_{\boldsymbol{x} \sim p_{\text{data }}(\boldsymbol{x})}[\log D(\boldsymbol{x} \mid \boldsymbol{y})] +  \notag \\ &  \mathbb{E}_{\boldsymbol{z} \sim p_z(\boldsymbol{z})}[\log (1-D(G(\boldsymbol{z} \mid \boldsymbol{y})))]
\end{align}
Here, $G$ and $D$ are the discriminator and generator respectively, and $x, y, z$ is the true sample, condition, and random seed separately. Given the latter should be helpful to stabilize training, we choose the loss in WGAN~\cite{arjovsky2017wasserstein}. Although it is just the non-log version Eq.~\ref{eq:loss1} (conditional case is Eq.~\ref{eq:loss2}) it mathmatically conforms to optimizing the best Wasserstein distance which effectively promotes the stableness of training.

\section{Experiments}
\subsection{Task \& Dataset}
Our objective is to synthesize images in either unconditional or conditional scenarios. As a preliminary experiment, we seek to validate the feasibility of CoC for generative tasks, for which we apply our approach to the MNIST~\cite{MNIST}, FashionMNIST~\cite{fmnist} and CryptoPunks 10k~\cite{crypto} dataset.

The MNIST dataset is a well-recognized collection, encompassing 60,000 training images and 10,000 testing images. These images, sized  $28\times28$  and single-channeled, represent handwritten digits. In the conditional scenario, the digit value serves as the condition guiding the generation process.

The FashionMNIST dataset is an MNIST-like dataset of 60,000 training images and 10,000 testing images, with a size of $28\times28$. Unlike MNIST dataset, the fashion MNIST dataset consists of fashionable images

The CryptoPunks 10k contains 10,000 images of the iconic CryptoPunks art collection, each uniquely generated using blockchain technology. The images showcase various characters with distinct features, including punk hairstyles, accessories, and facial expressions.

\subsection{Settings of Training}
The specifics of our model structure are detailed in Tab.~\ref{Tab:hyper} and we choose cluster center equal to $1$ for testing the performance of our model. Notably, $sample\_r$ denotes the rate of point increase or reduction in the upsampling or downsampling module, $heads$ and $head\_dim$ represent the number of heads in the CoC module and their corresponding dimensions, and $mlp\_r$ is the rate of dimension increase and reduction (to the same dimension before entering the modules) of the MLP, pertaining to transformation to and reduction from value space. For the WGAN version, the sigmoid activation in the discriminator is omitted.

To train the model, we employ the Adam optimizer~\cite{kingma2014adam} with a learning rate of $\mathbf{2e-4}$ and a batch size of 256. For the WGAN version, we utilize RMStrop~\cite{tieleman2012lecture} with identical learning rates and batch sizes. The cosine annealing decay policy is implemented to adjust the learning rate, and gradient cut-offs are used in the WGAN version. In all trials, the loss converges within 50 training epochs. Consequently, in addition to reviewing generated samples, we compute the best $FID$\cite{heusel2018gans}, $KID$\cite{bińkowski2021demystifying} and $IS$ \cite{salimans2016improved}to evaluate the performance of CoC-GAN.

\begin{figure}[!htbp]
    \includegraphics[width=0.35
    \linewidth]{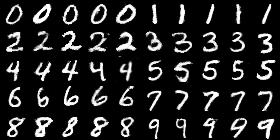}
    \includegraphics[width=0.35
    \linewidth]{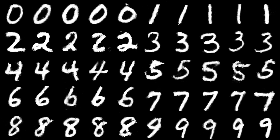}
    \caption{Results on MNIST of unconditional case (the left one) and conditional case (the right one).}
    \label{fig:unMNIST}
\end{figure}

\subsection{Generative Results on MNIST}
We present the performance results for the unconditional and conditional scenario in Tab.~\ref{tab:result}, with corresponding generated samples displayed in Fig.~\ref{fig:unMNIST}. It is evident that the implementation of the WGAN version significantly enhances the performance of the CoC-GAN. Intuitively, the CoC-GAN seems well-suited to generative tasks, albeit requiring further refinement in model design and training settings, such as the incorporation of WGAN. Furthermore, these results also validate the practicality of the Point Increaser module in incrementally generating points.~\label{sec:uncond}

\begin{figure}
    \includegraphics[width=0.35
    \linewidth]{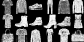}
    \includegraphics[width=0.35
    \linewidth]{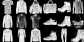}
    \caption{Results on FMNIST of unconditional case (the left one) and conditional case (the right one).}
    \label{fig:FMNIST}
\end{figure}

\subsection{Generative Results on FashionMNIST}
Besides conducting experiments on MNIST dataset, we also test our CoC-GAN in FashionMNIST to verify its ability for adapting different image sources. See concrete outcomes in Tab.~\ref{tab:result}. As results show, our CoC-GAN achieved satisfactory performance. Corresponding generated samples displayed in Fig.~\ref{fig:FMNIST}.

\subsection{Generative Results on 10K CryptoPunks}
\begin{figure}
    \includegraphics[width=0.65\linewidth]{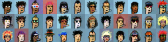}
   
    \caption{Results on 10K CryptoPunks of unconditional case.}
    \label{fig:CryptoPunks}
\end{figure}

The performance results for the unconditional scenario are displayed in Tab.~\ref{tab:result}, with corresponding generated samples presented in Fig.~\ref{fig:CryptoPunks}. These results reinforce the conclusion reached in Sec.~\ref{sec:uncond} and further emphasize the complexity and instability inherent to training the CoC structure. A comparison with the unconditional scenario reveals that incorporating a condition into the generation process presents unexpected challenges.
\subsection{The Interpretability of CoC-GAN}
\begin{figure}[!htbp]
    \centering
    \includegraphics[width=0.5\linewidth]{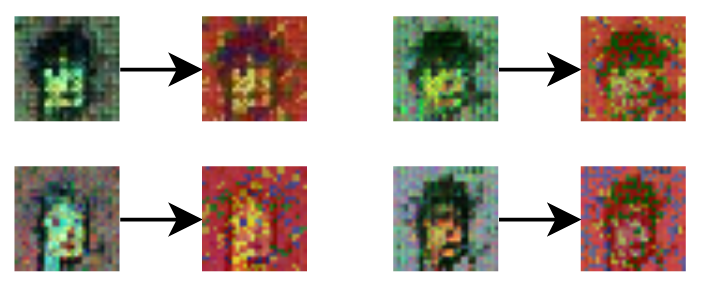}
    \caption{Illustration of visualization of 4 random examples. In each pair, the left is the original generative result, the right is the colored explanation for visualized clustering.}
    \label{fig:interpret}
\end{figure}

Considering the innate clustering interpretability of CoC-GAN, we set the centers of the first layer of its discriminator to 4 and retrain it on the 10K CryptoPunks dataset to demonstrate this property. Fig.~\ref{fig:interpret} visualizes how CoC-GAN performs its clustering work in the first layer of its discriminator. As shown, we can primarily deduce that the head part of the generative images is clustered into one category (visualized in yellow), and the background part is clustered into another category (visualized in red). This visualization illustrates that the clustering nature of CoC-GAN can provide substantial interpretability.

\section{Conclusion} 
In this paper, we introduce CoC-GAN, a novel approach that leverages the pure context clustering technique in both the generator and discriminator. In contrast to the point reducer used in downsampling processes, we propose the point increaser module to cater to the demand for generating an increasing number of points. Through empirical experiments conducted in both conditional and unconditional scenarios, we demonstrate that the combination of context clustering with the point increaser module can successfully tackle the generation task, albeit with careful architecture design and training settings. It is important to note that training the context clustering structure presents additional challenges compared to conventional MLP and convolution-based methods. While the performance of CoC-GAN may not surpass mainstream approaches, this study represents the first attempt to apply context clustering in the context of image generation. Building upon the feasibility established in this work, future research can explore the interpretability advantages of the context clustering technique and optimize its efficiency further.

\bibliographystyle{ACM-Reference-Format}
\bibliography{all.bib}


\begin{thebibliography}{22}


\ifx \showCODEN    \undefined \def \showCODEN     #1{\unskip}     \fi
\ifx \showDOI      \undefined \def \showDOI       #1{#1}\fi
\ifx \showISBNx    \undefined \def \showISBNx     #1{\unskip}     \fi
\ifx \showISBNxiii \undefined \def \showISBNxiii  #1{\unskip}     \fi
\ifx \showISSN     \undefined \def \showISSN      #1{\unskip}     \fi
\ifx \showLCCN     \undefined \def \showLCCN      #1{\unskip}     \fi
\ifx \shownote     \undefined \def \shownote      #1{#1}          \fi
\ifx \showarticletitle \undefined \def \showarticletitle #1{#1}   \fi
\ifx \showURL      \undefined \def \showURL       {\relax}        \fi
\providecommand\bibfield[2]{#2}
\providecommand\bibinfo[2]{#2}
\providecommand\natexlab[1]{#1}
\providecommand\showeprint[2][]{arXiv:#2}

\bibitem[Abdullah(2023)]%
        {crypto}
\bibfield{author}{\bibinfo{person}{Wasiq Abdullah}.}
  \bibinfo{year}{2023}\natexlab{}.
\newblock \bibinfo{title}{Pixelated Treasures: 10K CryptoPunks}.
\newblock
\newblock
\urldef\tempurl%
\url{https://doi.org/10.34740/KAGGLE/DSV/5061270}
\showDOI{\tempurl}


\bibitem[Arjovsky et~al\mbox{.}(2017)]%
        {arjovsky2017wasserstein}
\bibfield{author}{\bibinfo{person}{Martin Arjovsky}, \bibinfo{person}{Soumith
  Chintala}, {and} \bibinfo{person}{Léon Bottou}.}
  \bibinfo{year}{2017}\natexlab{}.
\newblock \bibinfo{title}{Wasserstein GAN}.
\newblock
\newblock
\showeprint[arxiv]{1701.07875}~[stat.ML]


\bibitem[Bińkowski et~al\mbox{.}(2021)]%
        {bińkowski2021demystifying}
\bibfield{author}{\bibinfo{person}{Mikołaj Bińkowski},
  \bibinfo{person}{Danica~J. Sutherland}, \bibinfo{person}{Michael Arbel},
  {and} \bibinfo{person}{Arthur Gretton}.} \bibinfo{year}{2021}\natexlab{}.
\newblock \bibinfo{title}{Demystifying MMD GANs}.
\newblock
\newblock
\showeprint[arxiv]{1801.01401}~[stat.ML]


\bibitem[Deng(2012)]%
        {MNIST}
\bibfield{author}{\bibinfo{person}{Li Deng}.} \bibinfo{year}{2012}\natexlab{}.
\newblock \bibinfo{title}{The mnist database of handwritten digit images for
  machine learning research}.
\newblock , \bibinfo{numpages}{141--142}~pages.
\newblock


\bibitem[Dosovitskiy et~al\mbox{.}(2021)]%
        {vit}
\bibfield{author}{\bibinfo{person}{Alexey Dosovitskiy}, \bibinfo{person}{Lucas
  Beyer}, \bibinfo{person}{Alexander Kolesnikov}, \bibinfo{person}{Dirk
  Weissenborn}, \bibinfo{person}{Xiaohua Zhai}, \bibinfo{person}{Thomas
  Unterthiner}, \bibinfo{person}{Mostafa Dehghani}, \bibinfo{person}{Matthias
  Minderer}, \bibinfo{person}{Georg Heigold}, \bibinfo{person}{Sylvain Gelly},
  \bibinfo{person}{Jakob Uszkoreit}, {and} \bibinfo{person}{Neil Houlsby}.}
  \bibinfo{year}{2021}\natexlab{}.
\newblock \bibinfo{title}{An Image is Worth 16x16 Words: Transformers for Image
  Recognition at Scale}.
\newblock
\newblock
\urldef\tempurl%
\url{https://openreview.net/forum?id=YicbFdNTTy}
\showURL{%
\tempurl}


\bibitem[Goodfellow et~al\mbox{.}(2020)]%
        {GAN}
\bibfield{author}{\bibinfo{person}{Ian~J. Goodfellow}, \bibinfo{person}{Jean
  Pouget{-}Abadie}, \bibinfo{person}{Mehdi Mirza}, \bibinfo{person}{Bing Xu},
  \bibinfo{person}{David Warde{-}Farley}, \bibinfo{person}{Sherjil Ozair},
  \bibinfo{person}{Aaron~C. Courville}, {and} \bibinfo{person}{Yoshua Bengio}.}
  \bibinfo{year}{2020}\natexlab{}.
\newblock \bibinfo{title}{Generative adversarial networks}.
\newblock , \bibinfo{numpages}{139--144}~pages.
\newblock
\urldef\tempurl%
\url{https://doi.org/10.1145/3422622}
\showDOI{\tempurl}


\bibitem[Han et~al\mbox{.}(2022)]%
        {vig}
\bibfield{author}{\bibinfo{person}{Kai Han}, \bibinfo{person}{Yunhe Wang},
  \bibinfo{person}{Jianyuan Guo}, \bibinfo{person}{Yehui Tang}, {and}
  \bibinfo{person}{Enhua Wu}.} \bibinfo{year}{2022}\natexlab{}.
\newblock \bibinfo{title}{Vision {GNN:} An Image is Worth Graph of Nodes}.
\newblock
\newblock
\urldef\tempurl%
\url{https://doi.org/10.48550/arXiv.2206.00272}
\showDOI{\tempurl}
\showeprint[arXiv]{2206.00272}


\bibitem[He et~al\mbox{.}(2016)]%
        {resnet}
\bibfield{author}{\bibinfo{person}{Kaiming He}, \bibinfo{person}{Xiangyu
  Zhang}, \bibinfo{person}{Shaoqing Ren}, {and} \bibinfo{person}{Jian Sun}.}
  \bibinfo{year}{2016}\natexlab{}.
\newblock \bibinfo{title}{Deep Residual Learning for Image Recognition}.
\newblock , \bibinfo{numpages}{770--778}~pages.
\newblock
\urldef\tempurl%
\url{https://doi.org/10.1109/CVPR.2016.90}
\showDOI{\tempurl}


\bibitem[Heusel et~al\mbox{.}(2018)]%
        {heusel2018gans}
\bibfield{author}{\bibinfo{person}{Martin Heusel}, \bibinfo{person}{Hubert
  Ramsauer}, \bibinfo{person}{Thomas Unterthiner}, \bibinfo{person}{Bernhard
  Nessler}, {and} \bibinfo{person}{Sepp Hochreiter}.}
  \bibinfo{year}{2018}\natexlab{}.
\newblock \bibinfo{title}{GANs Trained by a Two Time-Scale Update Rule Converge
  to a Local Nash Equilibrium}.
\newblock
\newblock
\showeprint[arxiv]{1706.08500}~[cs.LG]


\bibitem[Jiang et~al\mbox{.}(2021)]%
        {gransgan}
\bibfield{author}{\bibinfo{person}{Yifan Jiang}, \bibinfo{person}{Shiyu Chang},
  {and} \bibinfo{person}{Zhangyang Wang}.} \bibinfo{year}{2021}\natexlab{}.
\newblock \bibinfo{title}{TransGAN: Two Pure Transformers Can Make One Strong
  GAN, and That Can Scale Up}.
\newblock , \bibinfo{numpages}{14745--14758}~pages.
\newblock
\urldef\tempurl%
\url{https://proceedings.neurips.cc/paper/2021/hash/7c220a2091c26a7f5e9f1cfb099511e3-Abstract.html}
\showURL{%
\tempurl}


\bibitem[Kang et~al\mbox{.}(2023)]%
        {gigagan}
\bibfield{author}{\bibinfo{person}{Minguk Kang}, \bibinfo{person}{Jun{-}Yan
  Zhu}, \bibinfo{person}{Richard Zhang}, \bibinfo{person}{Jaesik Park},
  \bibinfo{person}{Eli Shechtman}, \bibinfo{person}{Sylvain Paris}, {and}
  \bibinfo{person}{Taesung Park}.} \bibinfo{year}{2023}\natexlab{}.
\newblock \bibinfo{title}{Scaling up GANs for Text-to-Image Synthesis}.
\newblock
\newblock
\urldef\tempurl%
\url{https://doi.org/10.48550/arXiv.2303.05511}
\showDOI{\tempurl}
\showeprint[arXiv]{2303.05511}


\bibitem[Kingma and Ba(2014)]%
        {kingma2014adam}
\bibfield{author}{\bibinfo{person}{Diederik~P Kingma} {and}
  \bibinfo{person}{Jimmy Ba}.} \bibinfo{year}{2014}\natexlab{}.
\newblock \bibinfo{title}{Adam: A method for stochastic optimization}.
\newblock
\newblock


\bibitem[Krizhevsky et~al\mbox{.}(2012)]%
        {alexnet}
\bibfield{author}{\bibinfo{person}{Alex Krizhevsky}, \bibinfo{person}{Ilya
  Sutskever}, {and} \bibinfo{person}{Geoffrey~E. Hinton}.}
  \bibinfo{year}{2012}\natexlab{}.
\newblock \bibinfo{title}{ImageNet Classification with Deep Convolutional
  Neural Networks}.
\newblock , \bibinfo{numpages}{1106--1114}~pages.
\newblock
\urldef\tempurl%
\url{https://proceedings.neurips.cc/paper/2012/hash/c399862d3b9d6b76c8436e924a68c45b-Abstract.html}
\showURL{%
\tempurl}


\bibitem[Lecun et~al\mbox{.}(1998)]%
        {LeNet}
\bibfield{author}{\bibinfo{person}{Y. Lecun}, \bibinfo{person}{L. Bottou},
  \bibinfo{person}{Y. Bengio}, {and} \bibinfo{person}{P. Haffner}.}
  \bibinfo{year}{1998}\natexlab{}.
\newblock \bibinfo{title}{Gradient-based learning applied to document
  recognition}.
\newblock , \bibinfo{numpages}{2278-2324}~pages.
\newblock
\urldef\tempurl%
\url{https://doi.org/10.1109/5.726791}
\showDOI{\tempurl}


\bibitem[Liu et~al\mbox{.}(2021)]%
        {swintrans}
\bibfield{author}{\bibinfo{person}{Ze Liu}, \bibinfo{person}{Yutong Lin},
  \bibinfo{person}{Yue Cao}, \bibinfo{person}{Han Hu}, \bibinfo{person}{Yixuan
  Wei}, \bibinfo{person}{Zheng Zhang}, \bibinfo{person}{Stephen Lin}, {and}
  \bibinfo{person}{Baining Guo}.} \bibinfo{year}{2021}\natexlab{}.
\newblock \bibinfo{title}{Swin Transformer: Hierarchical Vision Transformer
  using Shifted Windows}.
\newblock , \bibinfo{numpages}{9992--10002}~pages.
\newblock
\urldef\tempurl%
\url{https://doi.org/10.1109/ICCV48922.2021.00986}
\showDOI{\tempurl}


\bibitem[Ma et~al\mbox{.}(2023)]%
        {CoC}
\bibfield{author}{\bibinfo{person}{Xu Ma}, \bibinfo{person}{Yuqian Zhou},
  \bibinfo{person}{Huan Wang}, \bibinfo{person}{Can Qin}, \bibinfo{person}{Bin
  Sun}, \bibinfo{person}{Chang Liu}, {and} \bibinfo{person}{Yun Fu}.}
  \bibinfo{year}{2023}\natexlab{}.
\newblock \bibinfo{title}{Image as Set of Points}.
\newblock
\newblock
\urldef\tempurl%
\url{https://doi.org/10.48550/arXiv.2303.01494}
\showDOI{\tempurl}
\showeprint[arXiv]{2303.01494}


\bibitem[Radford et~al\mbox{.}(2016)]%
        {dcgan}
\bibfield{author}{\bibinfo{person}{Alec Radford}, \bibinfo{person}{Luke Metz},
  {and} \bibinfo{person}{Soumith Chintala}.} \bibinfo{year}{2016}\natexlab{}.
\newblock \bibinfo{title}{Unsupervised Representation Learning with Deep
  Convolutional Generative Adversarial Networks}.
\newblock
\newblock
\urldef\tempurl%
\url{http://arxiv.org/abs/1511.06434}
\showURL{%
\tempurl}


\bibitem[Reed et~al\mbox{.}(2016)]%
        {GANINTCLS}
\bibfield{author}{\bibinfo{person}{Scott Reed}, \bibinfo{person}{Zeynep Akata},
  \bibinfo{person}{Xinchen Yan}, \bibinfo{person}{Lajanugen Logeswaran},
  \bibinfo{person}{Bernt Schiele}, {and} \bibinfo{person}{Honglak Lee}.}
  \bibinfo{year}{2016}\natexlab{}.
\newblock \bibinfo{title}{Generative adversarial text to image synthesis}.
\newblock , \bibinfo{numpages}{1060--1069}~pages.
\newblock


\bibitem[Salimans et~al\mbox{.}(2016)]%
        {salimans2016improved}
\bibfield{author}{\bibinfo{person}{Tim Salimans}, \bibinfo{person}{Ian
  Goodfellow}, \bibinfo{person}{Wojciech Zaremba}, \bibinfo{person}{Vicki
  Cheung}, \bibinfo{person}{Alec Radford}, {and} \bibinfo{person}{Xi Chen}.}
  \bibinfo{year}{2016}\natexlab{}.
\newblock \bibinfo{title}{Improved techniques for training gans}.
\newblock
\newblock


\bibitem[Tieleman et~al\mbox{.}(2012)]%
        {tieleman2012lecture}
\bibfield{author}{\bibinfo{person}{Tijmen Tieleman}, \bibinfo{person}{Geoffrey
  Hinton}, {et~al\mbox{.}}} \bibinfo{year}{2012}\natexlab{}.
\newblock \bibinfo{title}{Lecture 6.5-rmsprop: Divide the gradient by a running
  average of its recent magnitude}.
\newblock , \bibinfo{numpages}{26--31}~pages.
\newblock


\bibitem[Tolstikhin et~al\mbox{.}(2021)]%
        {mipmixer}
\bibfield{author}{\bibinfo{person}{Ilya~O. Tolstikhin}, \bibinfo{person}{Neil
  Houlsby}, \bibinfo{person}{Alexander Kolesnikov}, \bibinfo{person}{Lucas
  Beyer}, \bibinfo{person}{Xiaohua Zhai}, \bibinfo{person}{Thomas Unterthiner},
  \bibinfo{person}{Jessica Yung}, \bibinfo{person}{Andreas Steiner},
  \bibinfo{person}{Daniel Keysers}, \bibinfo{person}{Jakob Uszkoreit},
  \bibinfo{person}{Mario Lucic}, {and} \bibinfo{person}{Alexey Dosovitskiy}.}
  \bibinfo{year}{2021}\natexlab{}.
\newblock \bibinfo{title}{MLP-Mixer: An all-MLP Architecture for Vision}.
\newblock , \bibinfo{numpages}{24261--24272}~pages.
\newblock
\urldef\tempurl%
\url{https://proceedings.neurips.cc/paper/2021/hash/cba0a4ee5ccd02fda0fe3f9a3e7b89fe-Abstract.html}
\showURL{%
\tempurl}


\bibitem[Xiao et~al\mbox{.}(2017)]%
        {fmnist}
\bibfield{author}{\bibinfo{person}{Han Xiao}, \bibinfo{person}{Kashif Rasul},
  {and} \bibinfo{person}{Roland Vollgraf}.} \bibinfo{year}{2017}\natexlab{}.
\newblock \bibinfo{title}{Fashion-MNIST: a Novel Image Dataset for Benchmarking
  Machine Learning Algorithms}.
\newblock
\newblock
\showeprint[arXiv]{cs.LG/1708.07747}~[cs.LG]


\end{thebibliography}

\end{document}